\documentclass[pmlr,twocolumn,10pt]{jmlr}

\jmlrvolume{LEAVE UNSET}
\jmlryear{2021}
\jmlrsubmitted{LEAVE UNSET}
\jmlrpublished{LEAVE UNSET}
\jmlrworkshop{Machine Learning for Health (ML4H) 2021} 

\usepackage[load-configurations=version-1]{siunitx} 
\usepackage{graphicx}
\usepackage{booktabs}       
\usepackage{multirow}
\usepackage{graphics}
\usepackage{flafter}
\usepackage{listings}
\usepackage{xcolor}
\usepackage{adjustbox}
\usepackage[skip=0pt]{caption}

\definecolor{codegreen}{rgb}{0,0.6,0}
\definecolor{codegray}{rgb}{0.5,0.5,0.5}
\definecolor{codepurple}{rgb}{0.58,0,0.82}
\definecolor{backcolour}{rgb}{0.95,0.95,0.92}

\lstdefinestyle{mystyle}{
    backgroundcolor=\color{backcolour},   
    commentstyle=\color{codegreen},
    keywordstyle=\color{magenta},
    numberstyle=\tiny\color{codegray},
    stringstyle=\color{codepurple},
    basicstyle=\ttfamily\footnotesize,
    breakatwhitespace=false,         
    breaklines=true,                 
    captionpos=b,                    
    keepspaces=true,                 
    numbers=left,                    
    numbersep=5pt,                  
    showspaces=false,                
    showstringspaces=false,
    showtabs=false,                  
    tabsize=2
}

\graphicspath{ {./images/APTOS} }
\usepackage[flushleft]{threeparttable}

\title{How Transferable are Self-supervised Features in Medical Image Classification Tasks?}

\author{%
\Name{Tuan Truong} \Email{tuan.truong@bayer.com}\\
\addr Bayer AG, Germany
\AND
\Name{Sadegh Mohammadi} \Email{sadegh.mohammadi@bayer.com}\\
\addr Bayer AG, Germany
\AND
\Name{Matthias Lenga} \Email{matthias.lenga@bayer.com}\\
\addr Bayer AG, Germany
}
\begin{document}

\maketitle

\begin{abstract}
Transfer learning has become a standard practice to mitigate the lack of labeled data in medical classification tasks. Whereas finetuning a downstream task using supervised ImageNet pretrained features is straightforward and extensively investigated in many works, there is little study on the usefulness of self-supervised pretraining. 
This paper assesses the transferability of the most recent self-supervised ImageNet models, including SimCLR, SwAV, and DINO, on selected medical imaging classification tasks. The chosen tasks cover tumor detection in sentinel axillary lymph node images, diabetic retinopathy classification in fundus images, and multiple pathological condition classification in chest X-ray images. 
We demonstrate that self-supervised pretrained models yield richer embeddings than their supervised counterparts, benefiting downstream tasks for linear evaluation and finetuning. For example, at a critically small subset of the data with linear evaluation, we see an im    
provement up to 14.79\% in Kappa score in the diabetic retinopathy classification task, 5.4\% in AUC in the tumor classification task, 7.03\% AUC in the pneumonia detection, and 9.4\% in AUC in the detection of pathological conditions in chest X-ray. 
In addition, we introduce \emph{Dynamic Visual Meta-Embedding} (DVME) as an end-to-end transfer learning approach that fuses pretrained embeddings from multiple models. We show that the collective representation obtained by DVME leads to a significant improvement in the performance of selected tasks compared to using a single pretrained model approach and can be generalized to any combination of pretrained models.


\end{abstract}



\section{Introduction}

\subsection{Background and Motivation}
The scarcity of high-quality annotated data remains a notorious challenge in medical image analysis due to the high cost of acquiring expert annotations \citep{Castro_2020}. Transfer learning from large models pretrained in a supervised fashion on natural images such as ImageNet has become a \textit{de-facto} solution for 2D medical imaging tasks in low data regimes \citep{lam_2018, Bayramoglu_2016, Procedia_2018, yang2018}. Recently, self-supervised learning shows initial success in building large-scale Deep Learning based applications by leveraging unannotated data for pretraining \citep{zhou_models_2019, chen_self-supervised_2019, taleb_3d_2020, taleb_multimodal_2020, zhuang2019, bai2019, martel_divide-and-rule_2020, sowrirajan_moco-cxr_2021}. However, a bottleneck within self-supervised learning is the demanding requirement of computational resources to train compared to standard supervised learning \citep{chen_simple_2020, caron_unsupervised_2021, caron_emerging_2021, grill_bootstrap_2020}. For example, regarding training on ImageNet, SwAV \citep{caron_unsupervised_2021} uses the batch size of 4096 distributed on 64 GPUs and SimCLR \citep{chen_simple_2020} uses varying batch sizes between 256 and 8192 on 32-128 TPU cores. Even when the batch size is small, the author of SimCLR notes that the training time must be extended to provide more negative examples. In pretraining medical datasets,  \cite{azizi_big_2021} observe the best performance when using the batch size of 1024 on 64 cloud TPU cores to train SimCLR on a chest X-ray dataset. While transfer learning from supervised pretraining on a large labeled dataset such as ImageNet is widely studied \citep{raghu_transfusion_2019, ke_chextransfer_2021}, the transferability of models pretrained on ImageNet using self-supervised techniques requires further investigation.

This paper reflects on the effectiveness of transfer learning with self-supervised features. We evaluate the performance of four downstream classification tasks using ImageNet pretrained features obtained from supervised and self-supervised techniques. The four distinct tasks concern three modalities with varying data sizes and distributions. The first task is in the domain of digital pathology and aims to detect sentinel axillary lymph node metastases in hematoxylin and eosin (H\&E) stained patches extracted from whole-slide images. The second task concerns the severity classification of diabetic retinopathy from colored fundus images. The last two tasks are related to reading X-ray images. One involves identifying whether a patient is suffering from pneumonia and the other involves detecting multiple findings, such as pneumothorax, nodule or mass, opacity, and fracture. 
In particular, we consider low data regimes ranging from approximately 1\% to 10\% of the original dataset size for each task (\sectionref{experiment:dataset}). 
We evaluate pretrained features of three self-supervised approaches, SimCLR \citep{chen_simple_2020}, SwAV \citep{caron_unsupervised_2021}, and DINO \citep{caron_emerging_2021}, on aforementioned tasks by training a linear layer on top of frozen features. We find that DINO consistently outperforms other self-supervised techniques and the supervised baseline by a significant margin. 

Additionally, we propose \textit{Dynamic Visual Meta-Embeddings} (DVME) - a model-agnostic technique to combine multiple self-supervised pretrained features for downstream tasks. In natural language processing, it has been observed that different word embeddings work well for different tasks and that it is difficult to anticipate the usefulness of a given embedding technique for a certain task at hand. The usage of a meta-embedding mitigates this problem by constructing an ensemble of embedding sets to increase the lexical coverage of vocabulary which leads to improved performance on downstream tasks \citep{kiela2018dynamic}. Similarly, in vision tasks, we propose to concatenate multiple pretrained embeddings with self-attention for transfer learning. Concatenation expands the embedding space and yields richer representation while self-attention adapts the contribution of individual embedding to a specific downstream task. We show that DVME leads to a further increase in performance across all tasks compared to the best single self-supervised pretrained model baseline.

\subsection{Contributions}
Overall, the main contributions are as following:
\begin{itemize}
    \item 
    Across four distinct medical image classification tasks, we assess the quality of the embeddings obtained from different models which are pretrained on ImageNet using state-of-the-art self-supervised or supervised pretraining techniques. 
    \item We identify a single self-supervised model which consistently outperforms the other approaches on all selected downstream tasks. In particular, this effect is prominently observed in low data regimes.
    \item We propose Dynamic Visual Meta-Embeddings (DVME) to fully leverage the collective representations obtained from different self-supervised pretrained models. The representations obtained from the DVME model aggregation outperform all single model approaches on the selected downstream tasks.
\end{itemize}

\section{Related work}

\paragraph{Self-supervised learning in medical imaging} Two main self-supervised approaches in medical imaging are in the form of \textit{handcrafted pretext tasks} and \textit{contrastive learning}. Early applications design tailored pretext tasks to reconstruct images from transformed or distorted inputs \citep{zhou_models_2019, chen_self-supervised_2019, taleb_multimodal_2020, zhuang2019, bai2019, rekik_modeling_2019, martel_divide-and-rule_2020}. For example, Model Genesis \citep{zhou_models_2019} applies in-domain transfer learning to various classification and segmentation tasks on CT and X-ray images. The proposed architecture is an autoencoder that reconstructs images from four transformations: non-linear, local-shuffling, out-painting, and in-painting. The induced transformations are supposed to enable the encoder to learn features related to appearance, texture, and context. \cite{chen_self-supervised_2019} propose context restoration as a pretext task applied in three common medical use cases: plane detection on fetal 2D ultrasound images, abdominal organ localization on CT images, and brain tumor segmentation on MRI images. The proposed method generates distorted images with different spatial contexts while maintaining the same intensity distribution by repeatedly swapping two random patches in an input image. Through reconstruction, the model learns useful semantic features transferable in subsequent target classification and segmentation tasks.
In a different approach, \cite{taleb_multimodal_2020} propose the multimodal puzzle task, inspired from the Jigsaw puzzles, which facilitates rich representation learning from multiple medical image modalities. 
The limitation of handcrafted pretext tasks is that they are highly task- and domain-specific, and thus cannot generalize well to different tasks. Lately, contrastive learning-based techniques (see \sectionref{methods}) resolve this issue. 
\cite{sowrirajan_moco-cxr_2021} use MoCo \citep{he_momentum_2020} to pretrain on unlabeled CheXpert \citep{irvin2019chexpert} dataset and finetunes with labels on external Shenzhen Hospital X-ray dataset \citep{QIMS5132} to detect pleural effusion. \cite{dippel2021finegrained} extends a contrastive loss to a self-reconstruction task with attention mechanism on fundus images.


\vspace{-0.02\linewidth}
\paragraph{Transfer learning in medical imaging} Transfer learning with ImageNet pretrained features still incites debates over its actual benefits for downstream medical tasks \citep{raghu_transfusion_2019, ke_chextransfer_2021,he_rethinking_2018}. In a large data regime, \cite{raghu_transfusion_2019} show that lightweight models with random initialization can perform on par with large architectures pretrained on ImageNet such as ResNet-50 \citep{he2015deep} and Inception-v3 \citep{inception}. On the contrary, \cite{ke_chextransfer_2021} argue that ImageNet pretraining can significantly boost the performance with newer architectures such as DenseNet \citep{densenet} and EfficientNet \citep{efficientnet}. In low data regimes, however, transfer learning with self-supervised approaches has been found particularly helpful in recent works \citep{newell_how_2020, azizi_big_2021, chaves_evaluation_2021}. \cite{azizi_big_2021} perform transfer learning with SimCLR \citep{chen_simple_2020} on X-ray and dermatology datasets and show a significant gain compared to a supervised baseline. \cite{chaves_evaluation_2021} evaluate self-supervised models on multiple dermatology datasets and find the advantage of self-supervised pretraining when using low training data. Whereas prior works focus on a single self-supervised technique \citep{azizi_big_2021} and a unique modality, i.e., dermatology \citep{chaves_evaluation_2021}, our work extends the investigation by benchmarking various self-supervised approaches against the supervised baseline across a set of heterogeneous medical imaging tasks. Our primary goal is to compare the richness of feature embeddings of different self-supervised learning techniques pretrained on ImgaeNet in the scope of transfer learning on medical imaging classification tasks.

\vspace{-0.028\linewidth}

\section{Self-supervised Learning Techniques}
\label{methods}
\defcitealias{chen_simple_2020}{SimCLR}
\defcitealias{caron_unsupervised_2021}{SwAV}
\defcitealias{caron_emerging_2021}{DINO}

An important line of work in self-supervised learning is contrastive learning where the representation is learned by comparing the similarity between images. The output embeddings obtained from an encoder are either pulled closer (similar) or pushed away (dissimilar) in the embedding space. Most of the contrastive approaches are built on the notion of \emph{multi-instance level classification} \citep{dosovitskiy2015discriminative}, which considers each image as a unique class and the model learns to discriminate it from the rest of the images in the batch. SimCLR and SwAV can be categorized into the group of contrastive learning. A detailed review and taxonomy of contrastive learning can be found in \citep{le-khac_contrastive_2020}.  There is also another line of work which does not discriminate the instance but matches the output features with those from a momentum encoder. BYOL \cite{grill_bootstrap_2020} and DINO are examples from this line. The techniques of our focus in the paper are \citetalias{chen_simple_2020}, \citetalias{caron_unsupervised_2021}, and \citetalias{caron_emerging_2021}. 

\paragraph{\citetalias{chen_simple_2020}} \textit{Simple Framework for Contrastive Learning of Visual Representation} \citep{chen_simple_2020} maximizes the agreement of two views from the same image. The paper
proposes a set of transformations applied to input images to create positive and negative pairs. An encoder takes a transformed batch and forwards it to a projection head that maps images to an embedding space. A contrastive loss on top compares the embeddings to minimize the distance between similar (positive) embeddings. Finally, the projection head is discarded and the encoder can be transferred to downstream tasks.

\paragraph{\citetalias{caron_unsupervised_2021}} \textit{Swapping Assignments between multiple Views of the same image} \citep{caron_unsupervised_2021} also contrasts two image views but not in a direct, sample-based fashion as \citetalias{chen_simple_2020}. Instead, it compares the cluster to which each view belongs. If two views come from the same image, they should fall on the same cluster assignment and vice versa. \cite{caron_unsupervised_2021} show that this approach has an advantage over \citetalias{chen_simple_2020} in avoiding the need for large batch size and improving the convergence time. In comparison to a prior clustering-based self-supervised technique in \citep{caron_deep_2019}, the clustering assignment process is online, so that gradients can be backpropagated in a batch-wise manner.

\paragraph{\citetalias{caron_emerging_2021}} \textit{Knowledge distillation without labels} \citep{caron_emerging_2021} matches the output probability distributions of two image views obtained from two networks. This approach takes inspiration from Bootstrap Your Own Latent (BYOL) \citep{grill_bootstrap_2020} in the perspective of self distillation task and the architecture of Vision Transformer (ViT) \citep{dosovitskiy_image_2021} as the backbone. Instead of passing the views into the same network, DINO passes two transformations of an image into two networks, namely the student and teacher network. The loss compares the probability outputs of both networks and the student's parameters are updated via backpropagation while the teacher's parameters is updated via an exponential moving average of the student ones. In addition, compared to using convolutional architectures, \cite{caron_emerging_2021}'s study indicates that ViT-based DINO shows distinct properties in characterizing object boundaries and generates features that perform well using K-Nearest Neighbors without further finetuning in ImageNet classification task.
\vspace{-0.04\linewidth}
\section{Datasets}
The four datasets in our experiments are distinct in terms of modality, dataset size, and class distribution to partially reflect the heterogeneity of typical medical imaging tasks. We consider three common modalities in medical image analysis: digital pathology, fundus imaging, and X-ray. 

\label{datasets}
\paragraph{PatchCamelyon (PatchCam)} 
The dataset contains H\&E sentinel axillary lymph node patches extracted from the whole-slide image in the study at \citep{10.1001/jama.2017.14585, veeling2018rotation}. All of the slides are annotated by expert pathologists. If the center of a patch contains at least one pixel of tumor tissue, it will be positive. The data version we use is the curated one from Kaggle competition\footnote{\url{https://www.kaggle.com/c/histopathologic-cancer-detection}} that removes all the duplicated patches and comes with a default train/test split. The original train set consists of 220025 patches of size 96$\times$96 with binary labels indicating whether there is a tumor or not. For our training task, we randomly select a subset comprising 50000 images.
The official test set comprises 57486 images for which no labels are provided. Hence, for all performance evaluations on PatchCam, we submit our predictions to Kaggle.

\paragraph{APTOS} 
The dataset comprises colored fundus images of Diabetic Retinopathy (DR) patients obtained from diverse clinics with different camera setups. For each image, clinicians rate the severity with a score between 0 and 4, indicating No-DR, Mild, Moderate, Severe, and Proliferative DR, respectively. The dataset is part of the challenge held on Kaggle\footnote{\url{https://www.kaggle.com/c/aptos2019-blindness-detection}}. The train and test sets contain 3662 and 1928 images, respectively. Similar to PatchCam, we submit the inference of the test set to Kaggle to obtain the scores.

\paragraph{Pneumonina chest X-ray} 

The dataset contains chest X-ray images annotated by two expert radiologists. Each radiologist classifies each image into healthy and pneumonia. We obtain the dataset in Kaggle \footnote{\url{https://www.kaggle.com/paultimothymooney/chest-xray-pneumonia}} with a default train/test split of 5216/624 images. Each patient can have multiple images, which is considered for patient stratification during training and validation. Further description of the dataset and acquisition can be found at \citep{KERMANY20181122}. 
\paragraph{NIH chest X-ray}
The dataset consists of chest X-ray images provided by NIH Clinical Center\footnote{\url{https://nihcc.app.box.com/v/ChestXray-NIHCC}}. Three certified radiologists manually reviewed the images.  Each radiologist marks the presence of four medical conditions: pneumothorax, nodule or mass, opacity, and fracture. We use two subsets of the original NIH Chest X-ray dataset, which are referred to as validation and test set in the study at \citep{doi:10.1148/radiol.2019191293}. We use the first subset (2414 images) for training and the second subset (1962 images) for evaluation. Since there can be multiple findings per image, we exclude such cases in our experiment for simplicity. In addition, we also add a class called \textit{other} for when no mentioned conditions are found in the image. 

\vspace{-0.04\linewidth}
\section{Experimental Setup}
\label{experiment}

\subsection{Architecture} 
To assess the transferability of self-supervised compare to fully supervised features pretrained on ImageNet, we used ResNet-50 \cite{he2015deep}, as it is the backbone for state-of-the-art self-supervised approaches \citep{chen_simple_2020, he_momentum_2020, caron_unsupervised_2021}, including SwAV \citep{caron_unsupervised_2021} and SimCLR \citep{chen_simple_2020}. On the other hand, for DINO \citep{caron_emerging_2021}, the architecture is VisionTransformer (ViT) with patch size 8$\times$8. We rely on a well-established computer VIsion library for state-of-the-art Self-Supervised Learning (VISSL) \citep{goyal2021vissl} for SwAV and SimCLR. At the same time, for DINO, we used available pretrained weight on the Facebook research repository \footnote{\url{https://github.com/facebookresearch/dino}}.

\vspace{-0.03\linewidth}
\subsection{Hyperparameters and Augmentation}
\label{experiment:hyperparams}
All experiments use the Adam optimizer starting with a small learning rate between 1e-3 and 1e-4 and further reducing it when the validation loss does not improve consecutively over five epochs. As our study aims to compare different initializations and not outperform the best performance, we do not apply intensive augmentation techniques. Images with an original size larger than 224$\times$224 are resized into 256$\times$256, then cropped and applied further flipping or rotation depending on the nature of modality. For the PatchCam dataset, we apply directly flipping without resizing or cropping as the size of the images is less than 224$\times$224. 

\vspace{-0.03\linewidth}
\subsection{Dataset sizes and subtasks}
\label{experiment:dataset}
To fully assess the transferability of self-supervised features under various data regimes, we define three different subtasks, Small (S), Medium (M), and Full (F). \tableref{table:data-portion} reports detailed information regarding data splits for each dataset. We construct the subtasks with a five-fold cross-validation fashion. We randomly extract samples from the entire dataset, conduct the training and validation for each split, and repeat this process five times. Then, we select the best performing model on the validation set to the fixed training set. To avoid bias learning due to class imbalance, we maintain the number of samples per class balance. The only exception is the NIH Chest X-ray dataset; we used an oversampling strategy during the training procedure because of a significant class imbalance.

\begin{table}[hbtp]
  \setlength{\tabcolsep}{3pt}
  \floatconts
  {table:data-portion}
  {\caption{Number of samples for different subtasks}}
  {\begin{tabular}{c c|c|c c}
    \toprule
    Dataset & S& M& F & Test \\
    \midrule
    PatchCam & 500 & 5000 & 50000 & 57486 \\
    APTOS & 50 & 500 & 3662 & 1928\\
    Pneumonia Chest X-ray & 50 & 500 & 5216 & 624 \\
    NIH Chest X-ray & 20 & 200 & 2414 & 1962 \\
    \bottomrule
  \end{tabular}

    }
\end{table}
\vspace{-0.03\linewidth}

\subsection{Metrics}
\label{experiment:metrics}
The evaluation metric for the PatchCam, Pneumonia and NIH Chest X-ray is the area under the Receiver Operating Characteristic curve (AUC) while the Cohen Kappa score for APTOS. We submit the APTOS and PatchCam test set predictions to Kaggle and obtain two scores for the private and public leaderboards, which are evaluated on two different portions of the test set. We calculate the final score as the weighted average score of the private and public leaderboard. Precisely, the  final score is calculated as $s_{avg} = \alpha \times s_{private} + (1 - \alpha) \times s_{public}$ where $\alpha$ for PatchCam is 0.51 and for APTOS is 0.85. The value of $\alpha$ is the portion of the test set that Kaggle uses to calculate the private leaderboard score and varies depending on the competition.

\vspace{-0.03\linewidth}
\subsection{Linear performance  and finetuning}
\label{experiment:training}
We conduct comprehensive experiments on pretrained models on ImageNet by two experiments: \emph{linear evaluation}  on frozen features and  \emph{finetuning} with labels of downstream tasks. For SwAV and SimCLR and fully supervised pretrained models on ImageNet, we add the linear layer after the last average pooling layer in ResNet-50. At the same time, for DINO, we follow the implementation of  \cite{caron_emerging_2021} to add a linear layer after the concatenation of class tokens from the last four blocks in ViT.  In addition to linear evaluation we consider \emph{finetuning}, where  all layers of the pretrained base network as well as the final linear classifier are adapted on the downstream task at hand. 

\vspace{-0.03\linewidth}
\subsection{Linear evaluation with DVME}
\label{experiment:dvme}
Given a set of pretrained feature extractors, it may be difficult to anticipate which pretrained model to choose for a given downstream task at hand. 
This concern is also shared in natural language processing where there are multiple word embedding techniques trained on different domains, each having its own strengths depending on target tasks. Meta-embedding is an effective technique that takes a union over different word embeddings to tackle the out-of-vocabulary problem and fuse multi-modal information \citep{kiela2018dynamic}. Though there is no multi-modal information in our study, we hypothesize that the pretrained features from different techniques are sufficiently independent of one another and encode certain complementary information. Therefore, we propose Dynamic Visual Meta-Embedding (DVME) for vision tasks, aggregating information by concatenating the embeddings of SimCLR, SwAV, and DINO pretrained models.
The newly constructed embedding space improves the separability of image features through the complementary effect of each embedding component. We extract the embedding space from the last fully connected unit from SimCLR and SwAV with the dimension 2048. For DINO, we construct the embedding by concatenating the class token of the last four blocks results in the dimension of 1536. Then, we project each embedding into a fixed size of 512 and feed the concatenation of the resulting embedding into a self-attention module. Then, we concatenate the embedding space and project it to a fixed dimension of 512 to learn the importance of each embedding component for a specific downstream task.
The self-attention module is the same as in the Vision Transformer architecture  \citep{dosovitskiy_image_2021} except that the attention is learned across different components of the meta-embedding instead of image patches. \figureref{fig:attn-mech} shows a sketch of how self-attention is incorporated when fusing the pretrained features. 
The proposed DVME approach is not limited to SimCLR, SwAV, or DINO and can be used with other feature extractors. We provide a snippet of DVME implemented in PyTorch in Appendix \sectionref{dvme}.
\vspace{-0.01\linewidth}

\begin{figure}[htbp]
\floatconts
    {fig:attn-mech}
    {\caption{Dynamic Visual Meta-Embedding (DVME): The embeddings extracted from each pretrained model are projected to the same dimension and concatenated before feeding to the self-attention module.}}
    {\includegraphics[width=\linewidth]{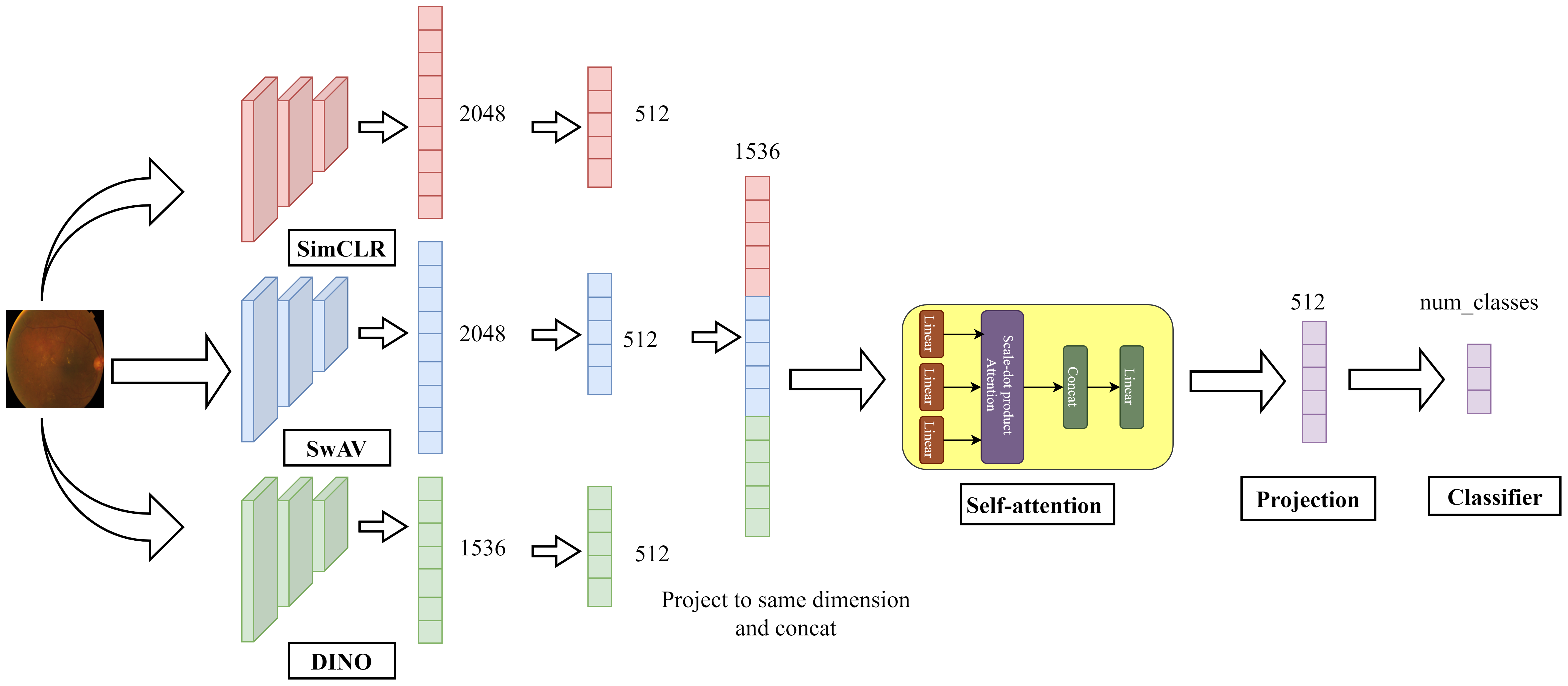}}
    \vspace{-0.04\linewidth}
\end{figure}


\section{Results and Discussion}
\label{result}
\subsection{Evaluation of self-supervised and supervised pretrained features}
\label{result-6.1}
\begin{figure*}[htbp]
\floatconts
     {fig:tsne}
     {\caption{t-SNE visualization of embeddings obtained using different pretrained feature extractors (supervised ImageNet, DINO, proposed method DVME). Top row: \textbf{PatchCam} dataset, bottom row: \textbf{APTOS} dataset}
     \vspace{-0.01\linewidth}
     }
     {
     \subfigure[Supervised ImageNet][t]{
         \label{fig:pcam-imagenet}
         \includegraphics[width=0.15\textwidth]{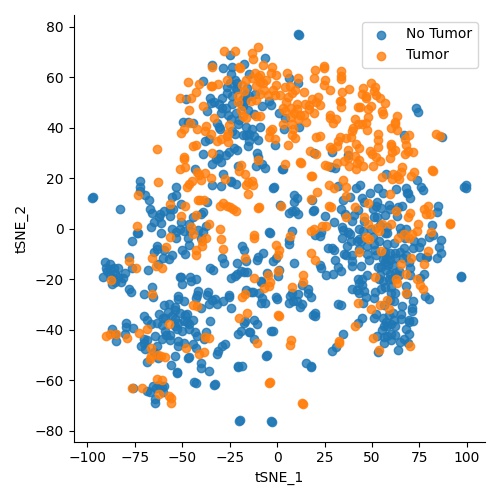}}
    \qquad
     \subfigure[DINO][t]{
         \label{fig:pcam-dino}
         \includegraphics[width=0.15\textwidth]{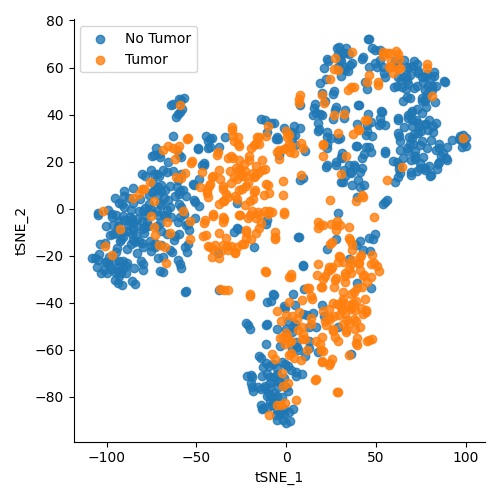}}
    \qquad
     \subfigure[DVME][t]{
         \label{fig:pcam-attn}
         \includegraphics[width=0.15\textwidth]{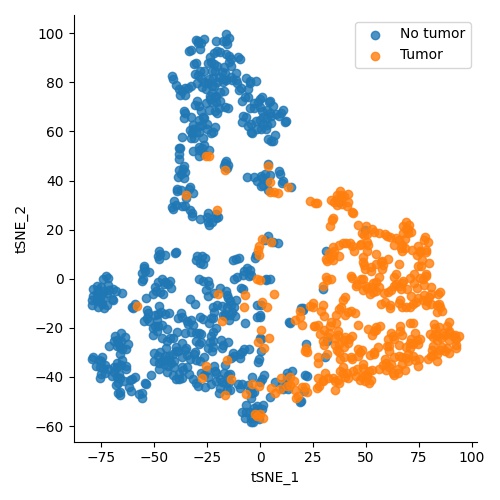}}
         
     \subfigure[Supervised ImageNet][t]{
         \label{fig:aptos-imagenet}
         \includegraphics[width=0.15\textwidth]{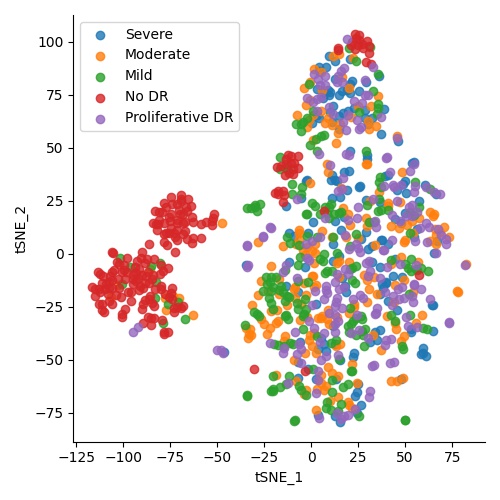}}
    \qquad
     \subfigure[DINO][t]{
         \label{fig:aptos-dino}
         \includegraphics[width=0.15\textwidth]{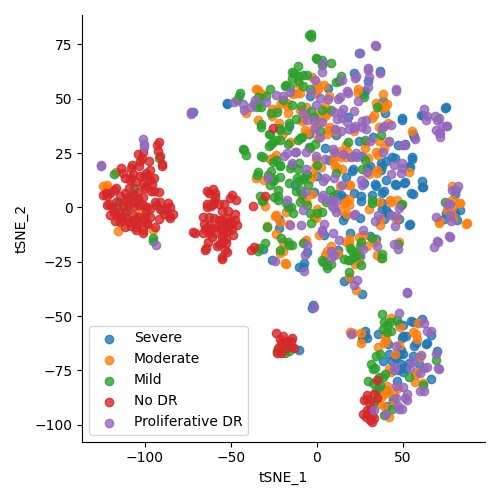}}
    \qquad
     \subfigure[DVME][t]{
         \label{fig:aptos-attn}
         \includegraphics[width=0.15\textwidth]{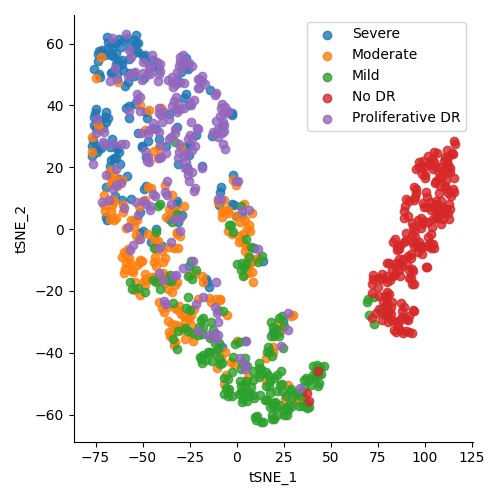}}
}
\end{figure*}

We test the generalization of self-supervised and fully supervised pretrained features on ImageNet by transferring them to several downstream medical imaging classification tasks under various data regimes. Table \ref{table:linear-eval} shows \emph{Linear evaluation} methods across various datasets. It is visible that SwAV and SimCLR pretrained features yield inconsistent patterns across all downstream tasks. For example, while SwAV and SimCLR initializations perform on par with each other on PatchCam and NIH Chest X-ray, they are different by approximately 10\% in Kappa score and 3.7\% in AUC on the S subsets of APTOS and Pneumonia Chest X-ray, respectively. Notably, DINO initialization consistently outperforms all the other initializations across all tasks by a significant margin.
For instance, on NIH Chest X-ray S and M subtasks, DINO pretrained features improve approximately 5-6\% in AUC over SimCLR and SwAV. 
The single exception is the APTOS S subtask, where SwAV outperforms DINO  by 3.3\% in AUC.
However, DINO still yields an improvement over SimCLR and ImageNet supervised initialization by 7\% and 11.2\% in Kappa score, respectively.
We refer to Appendix \ref{linear} for more detailed results on the performance obtained by the competing initialization methods for different dataset sizes. 
In comparison to ImageNet supervised pretrained features, we observe that self-supervised features improve the performance across all downstream tasks. This suggests that the representation generated by self-supervised methods are of higher quality, leading to better performance on the test set and reducing the performance variability between folds in low data regimes, similar to the observation made in \cite{chaves_evaluation_2021}.

\begin{table*}[hbtp]
  \setlength{\tabcolsep}{3pt}
  \floatconts
  {table:linear-eval}
  {\caption{Linear evaluation performance of different self-supervised initializations, supervised pretraining and random initialization on different scales (small, medium, full) of the data. }
  }
  {\begin{threeparttable}

  \begin{tabular}{c|c|ccc}
  
    \toprule
    Dataset & Method & Small (S) & Medium (M) & Full (F) \\
    \midrule
    \multirow{5}{5em}{PatchCam} & Random & 0.6594 $\pm$ 0.0319 & 0.6994 $\pm$ 0.0079 & 0.7990 $\pm$ 0.0021 \\
    & Supervised ImageNet & 0.7517 $\pm$ 0.0136 & 0.7863 $\pm$ 0.0063 & 0.7975 $\pm$ 0.0032  \\
    & SwAV   &  0.7834 $\pm$ 0.0112 & 0.8043 $\pm$ 0.0072 & 0.8088 $\pm$ 0.0025        \\
    & SimCLR & 0.7895 $\pm$ 0.0091 & 0.8053 $\pm$ 0.0069 & 0.8084 $\pm$ 0.0026  \\
    & DINO   & \bf{0.8058 $\pm$ 0.0100} & \bf{0.8359 $\pm$ 0.0053} & \bf{0.8487 $\pm$ 0.0014}  \\
    \midrule
    \multirow{5}{5em}{APTOS (*)} & Random & 0.0324 $\pm$ 0.0602 & 0.0624 $\pm$ 0.0459 & 0.1550 $\pm$ 0.1160  \\
    & Supervised ImageNet &  0.4851 $\pm$ 0.0811 & 0.6822 $\pm$ 0.0257 & 0.7331 $\pm$ 0.0124   \\
    & SwAV   &  \bf{0.6330 $\pm$ 0.0204} & 0.7274 $\pm$ 0.0095 & 0.7617 $\pm$ 0.0128       \\
    & SimCLR    & 0.5305 $\pm$ 0.0539 & 0.6500 $\pm$ 0.0138 & 0.6989 $\pm$ 0.0084  \\
    & DINO    & 0.6003 $\pm$ 0.0691 & \bf{0.7372 $\pm$ 0.0167} & \bf{0.7790 $\pm$ 0.0083} \\
    \midrule
    \multirow{5}{5em}{Pneumonia Chest X-ray} & Random & 0.6899 $\pm$ 0.0339 & 0.8258 $\pm$ 0.0237 & 0.8907 $\pm$ 0.0144 \\
    & Supervised ImageNet &  0.8789 $\pm$ 0.0234 & 0.8954 $\pm$ 0.0151 & 0.9397 $\pm$ 0.0033   \\
    & SwAV   &  0.8808 $\pm$ 0.0222 & 0.9215 $\pm$ 0.0252 & 0.9709 $\pm$ 0.0047       \\
    & SimCLR    & 0.9168 $\pm$ 0.0006 & 0.9346 $\pm$ 0.0072 & 0.9665 $\pm$ 0.0027  \\
    & DINO    & \bf{0.9492 $\pm$ 0.0170} & \bf{0.9718 $\pm$ 0.0055} & \bf{0.9868 $\pm$ 0.0008} \\
    \midrule
    \multirow{5}{5em}{NIH \\ Chest X-ray} &  Random & 0.5212 $\pm$ 0.0344  & 0.5317 $\pm$ 0.0176 & 0.5392 $\pm$ 0.0346\\
    & Supervised ImageNet &  0.5383 $\pm$ 0.0392 & 0.6688 $\pm$ 0.0148 & 0.7109 $\pm$ 0.0084   \\
    & SwAV   &  0.5785 $\pm$ 0.0258 & 0.6889 $\pm$ 0.0089 & 0.7225 $\pm$ 0.0139  \\
    & SimCLR    & 0.5792 $\pm$ 0.0435 & 0.6645 $\pm$ 0.0067 & 0.6983 $\pm$ 0.0231  \\
    & DINO & \bf{0.6323 $\pm$ 0.0131} & \bf{0.7373 $\pm$ 0.0112} & \bf{0.7438 $\pm$ 0.0228} \\
    \bottomrule
  \end{tabular}
  \begin{tablenotes}
      \small
      \item (*) The evaluation metric for APTOS is Cohen-Kappa score while for others is AUC score.
  \end{tablenotes}
  \end{threeparttable}}

\end{table*}

\begin{table*}[hbtp]
\vspace*{-0.15\linewidth}
\floatconts
{table:finetuning}
{\caption{Finetuning performance of different self-supervised initializations, supervised pretraining and random initialization on different scales (small, medium, full) of the data.}
}
  {\begin{threeparttable}

  \begin{tabular}{c|c|ccc}
    \toprule
    Dataset & Method & Small (S) & Medium (M) & Full (F) \\
    \midrule
    \multirow{5}{5em}{PatchCam} & Random & 0.7355 $\pm$ 0.0282 & 0.7660 $\pm$ 0.0223 & 0.8515 $\pm$ 0.0023 \\
    & Supervised ImageNet & 0.7897 $\pm$ 0.0162 & 0.8274 $\pm$ 0.0051 & 0.8483 $\pm$ 0.0097  \\
    & SwAV   &  0.7895 $\pm$ 0.0336 & 0.8399 $\pm$ 0.0142 & \bf{0.8619 $\pm$ 0.0090}        \\
    & SimCLR & 0.8021 $\pm$ 0.0138 & 0.8329 $\pm$ 0.0085 & 0.8553 $\pm$ 0.0110  \\
    & DINO   & \bf{0.8366 $\pm$ 0.0092} & \bf{0.8440 $\pm$ 0.0172} & 0.8517 $\pm$ 0.0158  \\
    \midrule
    \multirow{5}{5em}{APTOS (*)} & Random & 0.0177 $\pm$ 0.0954 & 0.3233 $\pm$ 0.0822 & 0.5927 $\pm$ 0.0545  \\
    & Supervised ImageNet &  0.4817 $\pm$ 0.0991 & 0.7369 $\pm$ 0.0310 & 0.8057 $\pm$ 0.0149   \\
    & SwAV   &  0.4928 $\pm$ 0.0378 & 0.7594 $\pm$ 0.0246 & 0.8293 $\pm$ 0.0133       \\
    & SimCLR    & 0.5916 $\pm$ 0.0570 & 0.7603 $\pm$ 0.0249 & 0.8264 $\pm$ 0.0103  \\
    & DINO    & \bf{0.6601 $\pm$ 0.0447} & \bf{0.7945 $\pm$ 0.0079} & \bf{0.8365 $\pm$ 0.0213} \\
    \midrule
    \multirow{5}{5em}{Pneumonia Chest X-ray} & Random & 0.6895 $\pm$ 0.0512 & 0.9183 $\pm$ 0.0186 & 0.9820 $\pm$ 0.0043 \\
    & Supervised ImageNet &  0.8649 $\pm$ 0.0442 & 0.9698 $\pm$ 0.0066 & 0.9910 $\pm$ 0.0015   \\
    & SwAV   &  \bf{0.9289 $\pm$ 0.0291} & 0.9814 $\pm$ 0.0087 & 0.9927 $\pm$ 0.0016       \\
    & SimCLR    & 0.9197 $\pm$ 0.0168 & 0.9781 $\pm$ 0.0085 & \bf{0.9950 $\pm$ 0.0013}  \\
    & DINO    & 0.9256 $\pm$ 0.0235 & \bf{0.9867 $\pm$ 0.0051} & 0.9948 $\pm$ 0.0010 \\
    \midrule
    \multirow{5}{5em}{NIH \\ Chest X-ray} &  Random & 0.5015 $\pm$ 0.0253  & 0.6404 $\pm$ 0.0165 & 0.6616 $\pm$ 0.0345\\
    & Supervised ImageNet &  0.5251 $\pm$ 0.0238 & 0.6816 $\pm$ 0.0429 & 0.7618 $\pm$ 0.0116   \\
    & SwAV   &  \bf{0.5903 $\pm$ 0.0384} & 0.6973 $\pm$ 0.0227 &\bf{ 0.7737 $\pm$ 0.0212}  \\
    & SimCLR    & 0.5570 $\pm$ 0.0450 & \bf{0.7228 $\pm$ 0.0287} & 0.7358 $\pm$ 0.0295  \\
    & DINO & 0.5552 $\pm$ 0.0546 & 0.6652 $\pm$ 0.0114 & 0.7404 $\pm$ 0.0240 \\
    \bottomrule
  \end{tabular}
  \begin{tablenotes}[para, flushleft]
      \small
      \item (*) The evaluation metric for APTOS is Cohen-Kappa score while for others is AUC score.
  \end{tablenotes}
  \end{threeparttable}
  \vspace*{-0.01\linewidth}
}
\end{table*}

\begin{table*}[htbp]
\vspace*{-0.2\linewidth}
\floatconts
{table:concat-emb}
{\caption{
  Linear evaluation performance of Dynamic Visual Meta-Embedding (DVME) in comparison with the best score obtained using a single pretrained model on different downstream tasks on different scales (small, medium, full) of the data.
  }}
  
  {
\begin{threeparttable}
  \begin{tabular}{c|c|ccc}
    \toprule
    Dataset     & Method & Small (S) & Medium (M) & Full (F)\\    
    \midrule
    \multirow{3}{7em}{PatchCam} 
    & DVEM   &  \bf{0.8227 $\pm$ 0.0148} & \bf{0.8399 $\pm$ 0.0059} & 0.8467 $\pm$ 0.0094\\
    & Best single baseline & 0.8058 $\pm$ 0.0100 & 0.8359 $\pm$ 0.0100  & \bf{0.8487 $\pm$ 0.0014} \\
    \midrule
    \multirow{3}{7em}{APTOS (*)} 
    & DVME   &  \bf{0.6913 $\pm$ 0.0575} & \bf{0.7925 $\pm$ 0.0265} & \bf{0.8242 $\pm$ 0.0279}\\
    & Best single baseline & 0.6330 $\pm$ 0.0204 & 0.7372 $\pm$ 0.0167 & 0.7790 $\pm$ 0.0083  \\
    \midrule
    \multirow{3}{7em}{Pneumonia Chest X-ray \\} 
    & DVME   &  \bf{0.9539 $\pm$ 0.0025} & 0.9696 $\pm$ 0.0101 & 0.9842 $\pm$ 0.0029\\
    & Best single baseline & 0.9492 $\pm$ 0.0170 & \bf{0.9718 $\pm$ 0.0055}  & \bf{0.9868 $\pm$ 0.0008}\\
    \midrule
    \multirow{3}{7em}{NIH \\Chest X-ray\\} 
    & DVME   &  \bf{0.6566 $\pm$ 0.0564} & \bf{0.7601 $\pm$ 0.0146} & \bf{0.7538 $\pm$ 0.0234}\\
    & Best single baseline & 0.6323 $\pm$ 0.0131 & 0.7373 $\pm$ 0.0112  & 0.7438 $\pm$ 0.0228\\
    \bottomrule
  \end{tabular}
    \begin{tablenotes}[para, flushleft]
      \small
      \item (*) The evaluation metric for APTOS is Cohen-Kappa score while for others is AUC score.
    \end{tablenotes}
\end{threeparttable}}
\vspace*{-4mm}
\end{table*}

 Figure \ref{fig:tsne} (a,b,d,e) shows the t-SNE visualization of the features extracted from the supervised pretrained ResNet-50 and DINO on the PatchCam (binary classification) and APTOS (multi-class) downstream tasks. It is visible that DINO offers a clear class separation compare to its supervised counterpart. We observe the same behavior for other datasets, which we refer to Appendix \ref{emb-visualize} for further detail.
 
We extend our comparison by \emph{finetuning} model initializations separately on all downstream tasks. 
Table \ref{table:finetuning} summarizes the finetuning results across datasets and their subtasks. Similar to the linear evaluation results, we consistently observe a higher performance for all self-supervised pretrained initializations compared to the supervised pretrained and randomly initialized baselines in the low data regimes (S, M), which supports the observation made by \cite{azizi_big_2021, chaves_evaluation_2021}.  DINO pretrained features outperform those from other self-supervised methods in 2/4 S subtasks and 3/4 M subtasks. When using full data  for fine-tuning, all self-supervised pretrained initializations consistently outperform the baseline methods on PatchCam, APTOS and Pneumonia Chest X-ray. However, for full NIH Chest X-ray task data, only SwAV exceeds the supervised baseline performance. 

Moreover, we observe that for S subtask of APTOS, Pneumonia Chest X-ray, and NIH Chest X-ray dataset where there are 50 samples or less finetuning leads to overfitting. For example, while DINO achieves the highest AUC of 0.6323 on the S subtask of NIH Chest X-ray in linear evaluation, the best performance for finetuning is obtained by SwAV and reaches only 0.5903 AUC. In the APTOS dataset, SwAV initialization reaches 0.6330 at the small dataset size in linear evaluation but drops to 0.4928 in finetuning. However, when the number of samples is increased up to a few thousand, the finetuning performance is higher than linear evaluation across all of the tasks. When using full data, the best performance using finetuning mounts up to 1.32\% in AUC, 5.75\% in Kappa score, 0.85\% in AUC, and 3\% in AUC in PatchCam, APTOS, Pneumonia Chest X-ray, and NIH Chest X-ray tasks, respectively. The improvement can be attributed to the increase in training samples, helping the model fit well to the downstream data but maintaining a decent generalization.

\vspace{-0.04\linewidth}
\subsection{Evaluation of DVME performance}
\label{result-6.2}
Table \ref{table:concat-emb} summarizes the results obtained from fusing SwAV, SimCLR, and DINO with the DVME approach. We evaluate DVME similar to the linear performance evaluation from Section \ref{result-6.1} by training only the meta-embedding on top of the three frozen feature extractors, cf. Section \ref{experiment:dvme}. As a performance benchmark, we select the self-supervised initialization for each dataset and each fraction of data that leads to the best linear evaluation performance, cf. Table \ref{table:linear-eval}.
DVME outperforms this benchmark in 4/4 of the S subtasks, 3/4 of the M subtasks, and 2/4 F subtasks. For the subtasks where DVME is not exceeding the benchmark performance, the difference lies within one standard deviation of the DVME linear evaluation score.
The improvement of DVME over the benchmark is particularly pronounced for the APTOS and NIH Chest X-ray tasks. For example, DVME helps gain roughly 6\% in Kappa score over the best individual baseline for the S and M subtask of the APTOS dataset.

The t-SNE visualizations of the DVME embeddings in Figure \ref{fig:tsne} (c,f) qualitatively indicate that the clusters are better separated, particularly in the case of multiclass classification. When analyzing the attention matrix, we find that SwAV and SimCLR pay little attention to each other but firmly into DINO, suggesting the representation from SwAV and SimCLR could be more similar and thus not so informative compared to DINO. 

To better understand the effect of self-attention on the model fusion, we conduct an ablation study on DVME in Appendix \ref{attn}. 
In the setting without self-attention, the meta-embedding is directly connected to the linear classifier. 
Without self-attention, the feature fusion still yields a significant improvement over the baseline, which supports our hypothesis in Section \ref{experiment:dvme} that each embedding contains complementary information. However, self-attention is particularly beneficial to specific tasks. For example, on APTOS, the Kappa scores increase by 5.6\%, 4.4\%, and 4.8\% for the S, M, and F subtask, respectively.

\vspace{-0.04\linewidth}
\section{Conclusion}
%
%
This study assesses the quality of ImageNet self-supervised pretrained features in four selected medical image classification tasks.
We demonstrate that feature extractor that is pretrained using SwAV, SimCLR, or DINO consistently yield richer embeddings on the downstream tasks than a supervised pretrained baseline model. Among all self-supervised techniques, DINO outperforms the other methods on the majority of datasets and subtasks. Furthermore, we show that each pretrained model's representations encode complementary information that can be fused to yield even more meaningful features.
Therefore, we propose Dynamic Visual Meta-Embedding (DVME), a model-agnostic meta-embedding approach. 
Our experiments indicate that DVME outperforms the best single model baseline on all downstream tasks.
As a model-agnostic approach, DVME is not limited to SwAV, SimCLR, or DINO. With slight modifications, other models can be combined using DVME to generate enriched representations.

\bibliography{ref}

\appendix
\onecolumn

\section{Dataset splits}
\label{datadetails}
For each experiment on a single dataset, we split the dataset into 5 training and validation folds. Each training fold contains  the same number of samples per each class. An exception is the NIH dataset since the number of samples across classes is highly imbalanced. In this case, we continue sampling to maximize the number of samples per each class as much as possible and use oversampling during the training process to compensate for class imbalance. In section \ref{detailedresults}, we report the sample size in absolute values across all of our experiments.

\section{Detailed results}
\label{detailedresults}
\subsection{Linear Evaluation}
\label{linear}
\setcounter{table}{0}
\renewcommand{\thetable}{B\arabic{table}}
For all experiments in linear evaluation, we replace the last layer of the pretrained model with a new linear classifier and train only this layer. The minimum and maximum number of epochs that we train our models are 30 and 50 respectively. In addition, we set early stopping with the patience of 10 epochs. The initial learing rate is 0.001 and is reduced with a factor of 0.1 by the \verb|ReduceLROnPlateau| scheduler when the validation score does not improve for 5 epochs consecutively. The batch size of 64 is kept fixed across all experiments.

\begin{table*}[htbp]
  \caption{Linear evaluation on the PatchCam dataset with various initializations. The mean AUC is obtained across 5 folds.\\}
  \label{table:linear-patchcam}
    \centering
  \resizebox{\linewidth}{!}{%
  \begin{tabular}{c|ccccc}
    \toprule
    \multicolumn{6}{c}{Mean AUC} \\
    \midrule
     Number of samples & Random & Supervised ImageNet & SwAV & SimCLR & DINO   \\    
    \midrule
    50 & 0.5041 $\pm$ 0.0091 & 0.7193 $\pm$ 0.0199 & 0.7543 $\pm$ 0.0342 & 0.7502 $\pm$ 0.0376 & 0.7721  $\pm$ 0.0233\\
    100 & 0.5001 $\pm$ 0.0002 & 0.7287 $\pm$ 0.0218 & 0.7756 $\pm$ 0.0280 & 0.7714 $\pm$ 0.0387 & 0.8040 $\pm$ 0.0050 \\
    200 & 0.5629 $\pm$ 0.0310 & 0.7183 $\pm$ 0.0224 & 0.7510 $\pm$ 0.0345 & 0.7675 $\pm$ 0.0189 & 0.8010 $\pm$ 0.0107\\
    500 (S) & 0.6594 $\pm$ 0.0319 & 0.7517 $\pm$ 0.0136 & 0.7834 $\pm$ 0.0112 & 0.7895 $\pm$ 0.0091 & 0.8058 $\pm$ 0.0100 \\
    1000 & 0.6886 $\pm$ 0.0090 & 0.7667 $\pm$ 0.0087 & 0.7686 $\pm$ 0.0509 & 0.7981 $\pm$ 0.0024 & 0.8204 $\pm$ 0.0106\\
    2000 & 0.6955 $\pm$ 0.0168 & 0.7709 $\pm$ 0.0075 & 0.8022 $\pm$ 0.0071 & 0.7996 $\pm$ 0.0076 & 0.8214 $\pm$ 0.0116\\
    5000 (M) & 0.6994 $\pm$ 0.0079 & 0.7863 $\pm$ 0.0063 & 0.8043 $\pm$ 0.0072 & 0.8053 $\pm$ 0.0069 & 0.8359 $\pm$ 0.0053\\
    10000 & 0.7110 $\pm$ 0.0046 & 0.7894 $\pm$ 0.0046 & 0.8338 $\pm$ 0.0163 & 0.8051 $\pm$ 0.0050 & 0.8399 $\pm$ 0.0029\\
    20000 & 0.7210 $\pm$ 0.0081 & 0.7970 $\pm$ 0.0061 & 0.8310 $\pm$ 0.0356 & 0.8110 $\pm$ 0.0048 & 0.8446 $\pm$ 0.0033 \\
    Full & 0.7990 $\pm$ 0.0021 & 0.7975 $\pm$ 0.0032 & 0.8088 $\pm$ 0.0025 & 0.8084 $\pm$ 0.0026 & 0.8487 $\pm$ 0.0014 \\

    \bottomrule
  \end{tabular}}
\end{table*}

\begin{table*}[htbp]
  \caption{Linear evaluation on the APTOS dataset with various initializations. The mean Kappa score is obtained across 5 folds.\\}
  \label{table:linear-aptos}
  \centering
  
  \resizebox{\linewidth}{!}{%
  \begin{tabular}{c|ccccc}
    \toprule
    \multicolumn{6}{c}{Mean Kappa Score} \\
    \midrule
     Number of samples & Random & Supervised ImageNet & SwAV & SimCLR & DINO   \\
    \midrule
    50 (S) & 0.0324 $\pm$ 0.0602  & 0.4851 $\pm$ 0.0811 & 0.6330 $\pm$ 0.0204 & 0.5305 $\pm$ 0.0539 & 0.6003 $\pm$ 0.0691 \\
    100    & -0.0272 $\pm$ 0.0300 & 0.5758 $\pm$ 0.0435 & 0.6559 $\pm$ 0.0260 & 0.5657 $\pm$ 0.0611 & 0.6889 $\pm$ 0.0433 \\
    200    & 0.0083 $\pm$ 0.0429  & 0.6752 $\pm$ 0.0219 & 0.7100 $\pm$ 0.0118 & 0.6369 $\pm$ 0.0084 & 0.7339 $\pm$ 0.0244\\
    500 (M)& 0.0624 $\pm$ 0.0459  & 0.6822 $\pm$ 0.0257 & 0.7274 $\pm$ 0.0095 & 0.6500 $\pm$ 0.0138 & 0.7372 $\pm$ 0.0167\\
    Full   & 0.1550 $\pm$ 0.1160  & 0.7331 $\pm$ 0.0124 & 0.7617 $\pm$ 0.0128 & 0.6989 $\pm$ 0.0084 & 0.7790 $\pm$ 0.0083\\

    \bottomrule
  \end{tabular}}
\end{table*}


\begin{table*}[htbp]
  \caption{Linear evaluation on the Pneumonia Chest X-Ray dataset with various initializations. The mean AUC is obtained across 5 folds.\\}
  \label{table:linear-xray}
  \centering
  \resizebox{\linewidth}{!}{%
  \begin{tabular}{c|ccccc}
    \toprule
    \multicolumn{6}{c}{Mean AUC} \\
    \midrule
     Number of samples & Random & Supervised ImageNet & SwAV & SimCLR & DINO   \\
    \midrule
    50 (S) & 0.6899 $\pm$ 0.0339 & 0.8789 $\pm$ 0.0234 & 0.8808 $\pm$ 0.0222 & 0.9168 $\pm$ 0.0006 & 0.9492 $\pm$ 0.0170 \\
    100    & 0.7323 $\pm$ 0.0602 & 0.8788 $\pm$ 0.0197 & 0.8731 $\pm$ 0.0260 & 0.9010 $\pm$ 0.0337 & 0.9466 $\pm$ 0.0154 \\
    200    & 0.7720 $\pm$ 0.0247 & 0.8789 $\pm$ 0.0315 & 0.8753 $\pm$ 0.0246 & 0.9176 $\pm$ 0.0154 & 0.9553 $\pm$ 0.0136 \\
    500 (M)& 0.8258 $\pm$ 0.0237 & 0.8954 $\pm$ 0.0151 & 0.9215 $\pm$ 0.0252 & 0.9346 $\pm$ 0.0072 & 0.9718 $\pm$ 0.0055\\
    Full   & 0.8907 $\pm$ 0.0144 & 0.9397 $\pm$ 0.0033 & 0.9709 $\pm$ 0.0047 & 0.9665 $\pm$ 0.0027 & 0.9868 $\pm$ 0.0008\\
    \bottomrule
  \end{tabular}}
\end{table*}


\begin{table*}[htbp]
  \caption{Linear evaluation on the NIH Chest X-ray dataset with various initializations. The mean AUC is obtained across 5 folds.\\}
  \label{table:linear-nih}
  \centering
  \resizebox{\linewidth}{!}{%
  \begin{tabular}{c|ccccc}
    \toprule
    \multicolumn{6}{c}{Mean AUC} \\
    \midrule
     Number of samples & Random & Supervised ImageNet & SwAV & SimCLR & DINO   \\
    \midrule
    20 (S) & 0.5212 $\pm$ 0.0344 & 0.5383 $\pm$ 0.0392 & 0.5785 $\pm$ 0.0258 & 0.5792 $\pm$ 0.0435 & 0.6323 $\pm$ 0.0131 \\
    50     & 0.5127 $\pm$ 0.0172 & 0.5897 $\pm$ 0.0283 & 0.6469 $\pm$ 0.0140 & 0.6273 $\pm$ 0.0130 & 0.6831 $\pm$ 0.0233 \\
    100    & 0.5031 $\pm$ 0.0465 & 0.6388 $\pm$ 0.0169 & 0.6563 $\pm$ 0.0238 & 0.6359 $\pm$ 0.0375 & 0.6686 $\pm$ 0.0227 \\
    150    & 0.5044 $\pm$ 0.0216 & 0.6432 $\pm$ 0.0283 & 0.6673 $\pm$ 0.0272 & 0.6686 $\pm$ 0.0143 & 0.7385 $\pm$ 0.0243\\
    200 (M)& 0.5317 $\pm$ 0.0176 & 0.6688 $\pm$ 0.0148 & 0.6889 $\pm$ 0.0089 & 0.6645 $\pm$ 0.0067 & 0.7373 $\pm$ 0.0112\\
    Full   & 0.5392 $\pm$ 0.0346 & 0.7109 $\pm$ 0.0084 & 0.7225 $\pm$ 0.0139 & 0.6983 $\pm$ 0.0231 & 0.7438 $\pm$ 0.0228\\

    \bottomrule
  \end{tabular}}
\end{table*}

\subsection{Finetuning}
\label{finetuning}
We keep all the hyperparameters the same as linear evaluation (Section \ref{linear}) when finetuning all models except DINO since it has a different architecture. For DINO, we start with a smaller learning rate of 0.0001 and use a smaller batch size of 16 instead.

\begin{table*}[htbp]
  \caption{Finetuning on the PatchCam dataset with various initializations. The mean AUC is obtained across 5 folds.\\}
  \label{table:finetuning-patchcam}
  \centering
  \resizebox{\linewidth}{!}{%
  \begin{tabular}{c|ccccc}
    \toprule
    \multicolumn{6}{c}{Mean AUC} \\
    \midrule
     Number of samples & Random & Supervised ImageNet & SwAV & SimCLR & DINO   \\
    \midrule
    50 & 0.5181 $\pm$ 0.0402 & 0.6359 $\pm$ 0.0863 & 0.6237 $\pm$ 0.1207 & 0.6631 $\pm$ 0.0578 & 0.7901 $\pm$ 0.0093\\
    100 & 0.5725 $\pm$ 0.0692 & 0.6844 $\pm$ 0.1123 & 0.6164 $\pm$ 0.0981 & 0.6116 $\pm$ 0.0630 & 0.8115 $\pm$ 0.0300\\
    200 & 0.7016 $\pm$ 0.0483 & 0.7656 $\pm$ 0.0395 & 0.6539 $\pm$ 0.0993 & 0.6305 $\pm$ 0.0845 & 0.8028 $\pm$ 0.0332\\
    500 (S) & 0.7355 $\pm$ 0.0282 & 0.7897 $\pm$ 0.0162 & 0.7895 $\pm$ 0.0336 & 0.8021 $\pm$ 0.0138 & 0.8366 $\pm$ 0.0092\\
    1000 & 0.7642 $\pm$ 0.0220 & 0.7870 $\pm$ 0.0433 & 0.8174  $\pm$ 0.0182 & 0.8160 $\pm$ 0.0096 & 0.8454 $\pm$ 0.0091\\
    2000 & 0.7674 $\pm$ 0.0118 & 0.7950 $\pm$ 0.0204 & 0.8221 $\pm$ 0.0255 & 0.7944 $\pm$ 0.0161 & 0.8438 $\pm$ 0.0220\\
    5000 (M) & 0.7660 $\pm$ 0.0223 & 0.8274 $\pm$ 0.0051 & 0.8399 $\pm$ 0.0142 & 0.8329 $\pm$ 0.0085 & 0.8440 $\pm$ 0.0172\\
    10000 & 0.7846 $\pm$ 0.0126 & 0.8338 $\pm$ 0.0079 & 0.8338 $\pm$ 0.0163 & 0.8402 $\pm$ 0.0118 & 0.8379 $\pm$ 0.0165\\
    20000 & 0.8114 $\pm$ 0.0110 & 0.8491 $\pm$ 0.0065 & 0.8587 $\pm$ 0.0116 & 0.8492 $\pm$ 0.0186 & 0.8745 $\pm$ 0.0045\\
    Full & 0.8515 $\pm$ 0.0023 & 0.8483 $\pm$ 0.0097 & 0.8619 $\pm$ 0.0090 & 0.8553 $\pm$ 0.0110 & 0.8517 $\pm$ 0.0158\\

    \bottomrule
  \end{tabular}}
\end{table*}


\begin{table*}[htbp]
  \caption{Finetuning on the APTOS dataset with various initializations. The mean Kappa score is obtained across 5 folds.\\}
  \label{table:finetuning-aptos}
  \centering
  \resizebox{\linewidth}{!}{%
  \begin{tabular}{c|ccccc}
    \toprule
    \multicolumn{6}{c}{Mean Kappa Score} \\
    \midrule
     Number of samples & Random & Supervised ImageNet & SwAV & SimCLR & DINO   \\
    \midrule
    50 (S) & 0.0177 $\pm$ 0.0954 & 0.4817 $\pm$ 0.0991 & 0.4928 $\pm$ 0.0378 & 0.5916 $\pm$ 0.0570 & 0.6601 $\pm$ 0.0447\\
    100    & 0.8080 $\pm$ 0.0759 & 0.5289 $\pm$ 0.0930 & 0.6015 $\pm$ 0.0784 & 0.6085 $\pm$ 0.0412 & 0.7144 $\pm$ 0.0709\\
    200    & 0.1914 $\pm$ 0.0445 & 0.6634 $\pm$ 0.0405 & 0.7354 $\pm$ 0.0120 & 0.6860 $\pm$ 0.0412 & 0.7754 $\pm$ 0.0194\\
    500 (M)& 0.3233 $\pm$ 0.0822 & 0.7369 $\pm$ 0.0310 & 0.7594 $\pm$ 0.0246 & 0.7603 $\pm$ 0.0249 & 0.7945 $\pm$ 0.0079\\
    Full   & 0.5927 $\pm$ 0.0545 & 0.8057 $\pm$ 0.0149 & 0.8293 $\pm$ 0.0133 & 0.8264 $\pm$ 0.0103 & 0.8365 $\pm$ 0.0213\\
    \bottomrule
  \end{tabular}}
\end{table*}

\begin{table*}[htbp]
  \caption{Finetuning on the Pneumonia Chest X-ray dataset with various initializations. The mean AUC is obtained across 5 folds.\\}
  \label{table:finetuning-xray}
  \centering
  \resizebox{\linewidth}{!}{%
  \begin{tabular}{c|ccccc}
    \toprule
    \multicolumn{6}{c}{Mean AUC} \\
    \midrule
     Number of samples & Random & Supervised ImageNet & SwAV & SimCLR & DINO   \\
    \midrule
    50 (S) & 0.6895 $\pm$ 0.0512 & 0.8649 $\pm$ 0.0442 &  0.9289 $\pm$ 0.0291 &  0.9197 $\pm$ 0.0168 &  0.9256 $\pm$ 0.0235\\
    100    & 0.8210 $\pm$ 0.0683 &  0.9032 $\pm$ 0.0423 &  0.9248 $\pm$ 0.0361 &  0.9199 $\pm$ 0.0338 &  0.9363 $\pm$ 0.0188\\
    200    & 0.9004 $\pm$ 0.0078 &  0.9157 $\pm$ 0.0167 &  0.9593 $\pm$ 0.0125 &  0.9436 $\pm$ 0.0194 &  0.9687 $\pm$ 0.0086\\
    500 (M)& 0.9183 $\pm$ 0.0186 & 0.9698 $\pm$ 0.0066 & 0.9814 $\pm$ 0.0087 &  0.9781 $\pm$ 0.0085 &  0.9867 $\pm$ 0.0051\\
    Full   & 0.9820 $\pm$ 0.0043 &  0.9910 $\pm$ 0.0015 &  0.9927 $\pm$ 0.0016 &  0.9950 $\pm$ 0.0013 &  0.9948 $\pm$ 0.0010\\
    \bottomrule
  \end{tabular}}
\end{table*}

\begin{table*}[htbp]
  \caption{Finetuning on the NIH Chest X-ray with various initializations. The mean AUC is obtained across 5 folds.\\}
  \label{table:finetuning-nih}
  \centering
  \resizebox{\linewidth}{!}{%
  \begin{tabular}{c|ccccc}
    \toprule
    \multicolumn{6}{c}{Mean AUC} \\
    \midrule
     Number of samples & Random & Supervised ImageNet & SwAV & SimCLR & DINO   \\
    \midrule
    20 (S) & 0.5015 $\pm$ 0.0253 & 0.5251 $\pm$ 0.0238 & 0.5903 $\pm$ 0.0384 & 0.5570 $\pm$ 0.0450 & 0.5552 $\pm$ 0.0546\\
    50     & 0.5492 $\pm$ 0.0828 & 0.6105 $\pm$ 0.0381 & 0.6172 $\pm$ 0.0167 & 0.6227 $\pm$ 0.0309 & 0.6348 $\pm$ 0.0286\\
    100    & 0.5961 $\pm$ 0.0602 & 0.6567 $\pm$ 0.0357 & 0.6616 $\pm$ 0.0436 & 0.6768 $\pm$ 0.0773 & 0.6689 $\pm$ 0.0240\\
    150    & 0.6114 $\pm$ 0.0293 & 0.6639 $\pm$ 0.0347 & 0.7037 $\pm$ 0.0558 & 0.6795 $\pm$ 0.0515 & 0.6551 $\pm$ 0.0318\\
    200 (M)& 0.6404 $\pm$ 0.0165 & 0.6816 $\pm$ 0.0429 & 0.6973 $\pm$ 0.0227 & 0.7228 $\pm$ 0.0287 & 0.6652 $\pm$ 0.0114\\
    Full   & 0.6616 $\pm$ 0.0345 & 0.7618 $\pm$ 0.0116 & 0.7737 $\pm$ 0.0212 & 0.7358 $\pm$ 0.0295 & 0.7404 $\pm$ 0.0240\\
    \bottomrule
  \end{tabular}}
\end{table*}

\newpage
\section{Detailed results of DVME}
\label{attn}
Embedding from each self-supervised pretrained model is projected into a dimension of 512. Embeddings from SimCLR, SwAV, and DINO add up to a dimension of 1536. The self-attention module is implemented based on the timm libarary\footnote{\url{https://github.com/rwightman/pytorch-image-models/blob/master/timm/models/vision_transformer.py
}}. The output from the self-attention layer is further projected to a 512-dimensional layer followed by a ReLU layer, and the final linear layer. Table \ref{table:attn-patchcam}-\ref{table:attn-nih} show the detailed result of linear evaluation using DVME with and without self-attention.

\setcounter{table}{0}
\renewcommand{\thetable}{C\arabic{table}}
\begin{table*}[htbp]
  \caption{Linear evaluation using DVME on the PatchCam dataset. The mean AUC is obtained across 5 folds.\\}
  \label{table:attn-patchcam}
  \centering
  \begin{tabular}{c|cc}
    \toprule
    \multicolumn{3}{c}{Mean AUC} \\
    \midrule
     Number of samples & DVME w/o self-attention & DVME \\
    \midrule
    50 & 0.7376 $\pm$ 0.0350 & 0.7456 $\pm$ 0.0467 \\
    100 & 0.7906 $\pm$ 0.0226 & 0.7864 $\pm$ 0.0405 \\
    200 & 0.8076 $\pm$ 0.0182 & 0.8026 $\pm$ 0.0209\\
    500 (S) & 0.8196 $\pm$ 0.0100 & 0.8227 $\pm$ 0.0148 \\
    1000 & 0.8200 $\pm$ 0.0045 & 0.8316 $\pm$ 0.0112 \\
    2000 & 0.8242 $\pm$ 0.0083 & 0.8243 $\pm$ 0.0184\\
    5000 (M) & 0.8442 $\pm$ 0.0074 & 0.8399  $\pm$ 0.0059\\
    10000 & 0.8417 $\pm$ 0.0044 & 0.8404 $\pm$ 0.0068\\
    20000 & 0.8525 $\pm$ 0.0049 & 0.8444 $\pm$ 0.0100\\
    Full & 0.8478 $\pm$ 0.0052 & 0.8467 $\pm$ 0.0094\\

    \bottomrule
  \end{tabular}
\end{table*}

\begin{table*}[htbp]
  \caption{Linear evaluation using DVME on the APTOS dataset. The mean Kappa score is obtained across 5 folds.\\}
  \label{table:attn-aptos}
  \centering

  \begin{tabular}{c|cc}
    \toprule
    \multicolumn{3}{c}{Mean Kappa Score} \\
    \midrule
     Number of samples & DVME w/o self-attention & DVME \\
    \midrule
    50 (S) & 0.6354 $\pm$ 0.0428 & 0.6913 $\pm$ 0.0575 \\
    100    & 0.7018 $\pm$ 0.0175 & 0.6992 $\pm$ 0.0860 \\
    200    & 0.7351 $\pm$ 0.0240 & 0.7787 $\pm$ 0.0191 \\
    500 (M)& 0.7681 $\pm$ 0.0166 & 0.7925 $\pm$ 0.0265 \\
    Full   & 0.7759 $\pm$ 0.0134 & 0.8242 $\pm$ 0.0279 \\
    \bottomrule
  \end{tabular}
\end{table*}

\begin{table*}[htbp]
  \caption{Linear evaluation using DVME on the Pneumonia Chest X-ray dataset. The mean AUC is obtained across 5 folds.\\}
  \label{table:attn-xray}
  \centering

  \begin{tabular}{c|cc}
    \toprule
    \multicolumn{3}{c}{Mean AUC} \\
    \midrule
     Number of samples & DVME w/o self-attention & DVME \\
    \midrule
    50 (S) & 0.9543 $\pm$ 0.0072 & 0.9539 $\pm$ 0.0025 \\
    100    & 0.9528 $\pm$ 0.0128 & 0.9469 $\pm$ 0.0111 \\
    200    & 0.9532 $\pm$ 0.0177 & 0.9569 $\pm$ 0.0170 \\
    500 (M)& 0.9725 $\pm$ 0.0030 & 0.9696 $\pm$ 0.0101 \\
    Full   & 0.9865 $\pm$ 0.0030 & 0.9842 $\pm$ 0.0029 \\
    \bottomrule
  \end{tabular}
\end{table*}


\begin{table*}[htbp]
  \caption{Linear evaluation using DVME on the NIH Chest X-ray dataset. The mean AUC is obtained across 5 folds.\\}
  \label{table:attn-nih}
  \centering
  \begin{tabular}{c|cc}
    \toprule
    \multicolumn{3}{c}{Mean AUC} \\
    \midrule
     Number of samples & DVME w/o self-attention & DVME \\
    \midrule
    20 (S) & 0.6525 $\pm$ 0.0558 & 0.6566 $\pm$ 0.0564\\
    50     & 0.7051 $\pm$ 0.0255 & 0.6871 $\pm$ 0.0400 \\
    100    & 0.7260 $\pm$ 0.0130 & 0.7091 $\pm$ 0.0428 \\
    150    & 0.7209 $\pm$ 0.0179 & 0.7437 $\pm$ 0.0310 \\
    200 (M)& 0.7232 $\pm$ 0.0267 & 0.7601 $\pm$ 0.0146 \\
    Full   & 0.7575 $\pm$ 0.0177 & 0.7538 $\pm$ 0.0234 \\
    \bottomrule
  \end{tabular}
\end{table*}
\newpage
\section{Embedding visualization}
\label{emb-visualize}
\setcounter{figure}{0}
\renewcommand{\thefigure}{D\arabic{figure}}

\begin{figure*}[htbp]

     \floatconts
     {fig:tsne-pneuxr}
     {\caption{t-SNE visualization of the pretrained embeddings from supervised ImageNet, DINO, and our proposed method DVME on \bf{Pneumonary Chest X-ray} dataset }}
     {
     \subfigure[Supervised ImageNet][b]{
         \label{fig:pneuxr-imagenet}
         \includegraphics[width=0.3\textwidth]{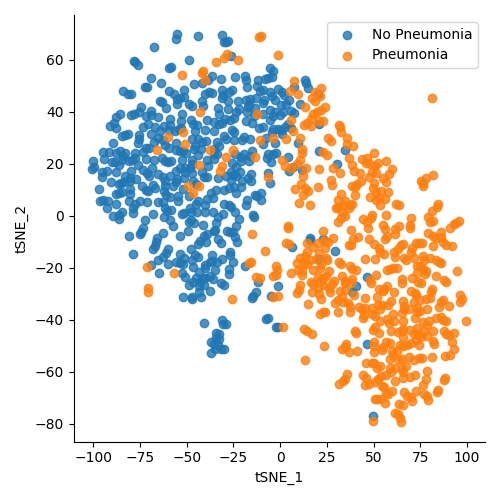}}
     \subfigure[DINO][b]{
        \label{fig:pneuxr-dino}
         \includegraphics[width=0.3\textwidth]{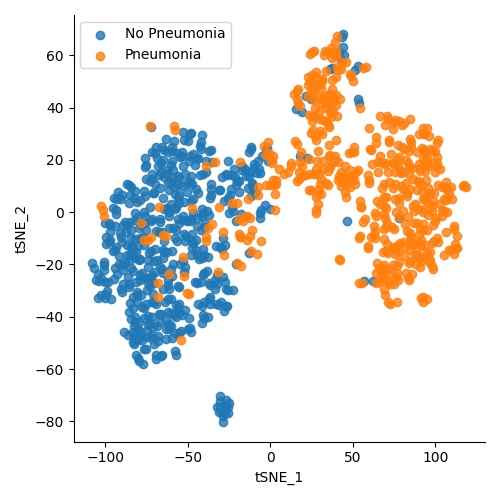}}
     \subfigure[DVME][b]{
         \label{fig:pneuxr-attn}
         \includegraphics[width=0.3\textwidth]{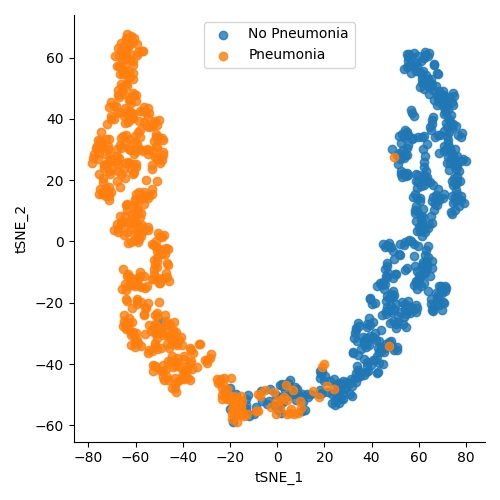}}
    }
\end{figure*}
\begin{figure*}[htbp]
    \floatconts
    {fig:tsne-nih}
    {\caption{t-SNE visualization of the pretrained embeddings from supervised ImageNet, DINO, and our proposed method DVME on \bf{NIH Chest X-ray} dataset }}
    {
    \subfigure[Supervised ImageNet][b]{
        \label{fig:nih-imagenet}
         \includegraphics[width=0.3\textwidth]{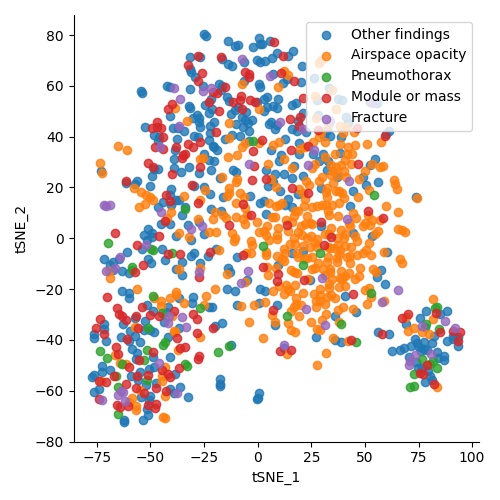}}
     \subfigure[DINO][b]{
        \label{fig:nih-dino}
         \includegraphics[width=0.3\textwidth]{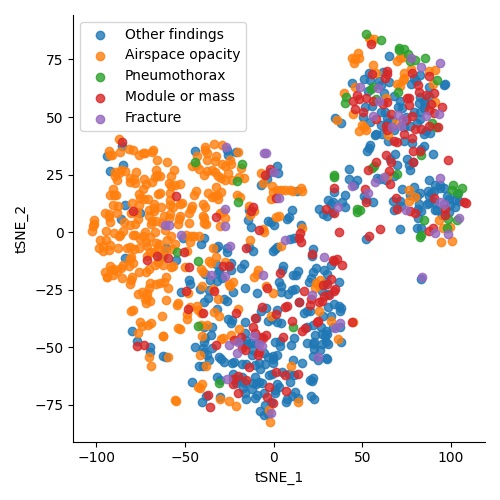}}
     \subfigure[DVME][b]{
        \label{fig:nih-attn}
         \includegraphics[width=0.3\textwidth]{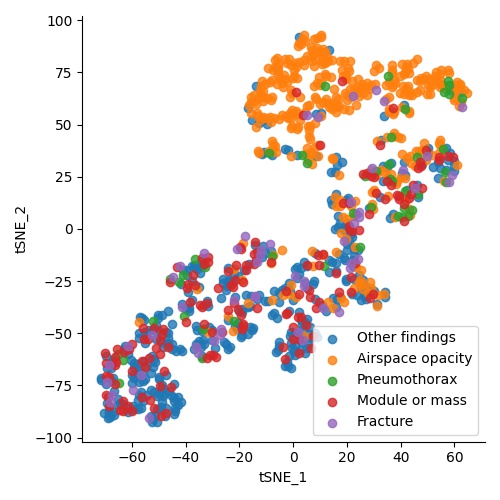}}
    }
\end{figure*}

\newpage
\section{Dynamic Visual Meta-embeddings}
\label{dvme}
\begin{lstlisting}[language=Python, caption=PyTorch code snippet of DVME, float]
import torch.nn as nn
import torch

class DVME(nn.Module):

    def __init__(self, proj_dim, num_cls, attn):
        # proj_dim: dimension of projection (default is 512)
        # num_cls:  number of classes
        # attn:     self-attention module
        
        super(DVME, self).__init__()
        self.simclr_head = nn.Linear(2048, proj_dim)
        self.swav_head = nn.Linear(2048, proj_dim)
        self.dino_head = nn.Linear(1536, proj_dim)
        self.attn = attn
        self.normlayer = nn.LayerNorm(proj_dim*3)
        self.proj_head = nn.Linear(proj_dim*3, proj_dim)
        self.classifier = nn.Linear(proj_dim, num_cls)
        self.dropout = nn.Dropout(0.2)
        
        
        
    def forward(self, x):
        # x:    dictionary containing extracted embeddings from 
        #       pretrained models SimCLR, SwAV, DINO
        
        simclr_out = self.simclr_head(x['simclr'])
        swav_out = self.swav_head(x['swav'])
        dino_out = self.dino_head(x['dino'])
        meta_x = torch.cat([simclr_out, swav_out, dino_out], dim=1)
        # reshape the meta-emb into (batch, tokens, dim)
        meta_x = meta_x.view(meta_x.size(0), -1, 1)
        out = self.attn(meta_x)
        out = self.normlayer(out.view(out.size(0), -1))
        out = self.proj_head(out).relu()
        out = self.dropout(out)
        out = self.classifier(out)
        return out
\end{lstlisting}

\newpage
\section{Attention weights}
\label{attn-weights}
\setcounter{figure}{0}
\renewcommand{\thefigure}{F\arabic{figure}}

\begin{figure*}[htbp]
    \floatconts
    {fig:attn-weights}
    {\caption{The self-attention weights obtained from the self-attention layers of DVME. At each dataset, the self-attention weights are averaged across all samples in test set using the models trained at the full size of dataset.}}
    {
    \subfigure[PatchCam][b]{
        \label{fig:attn-patchcam}
         \includegraphics[width=0.4\textwidth]{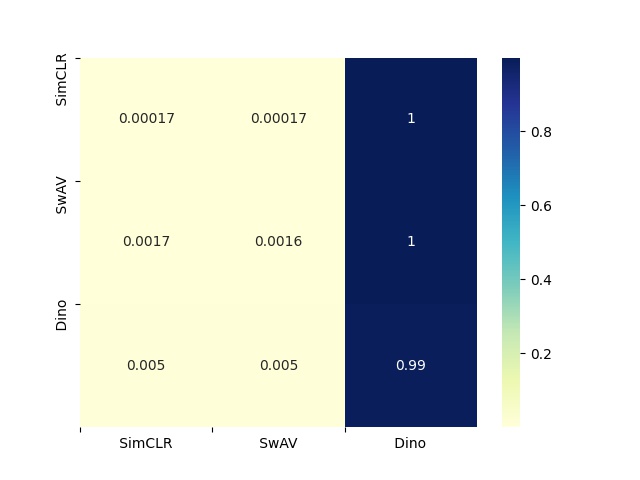}}
     \subfigure[APTOS][b]{
        \label{fig:attn-aptos}
         \includegraphics[width=0.4\textwidth]{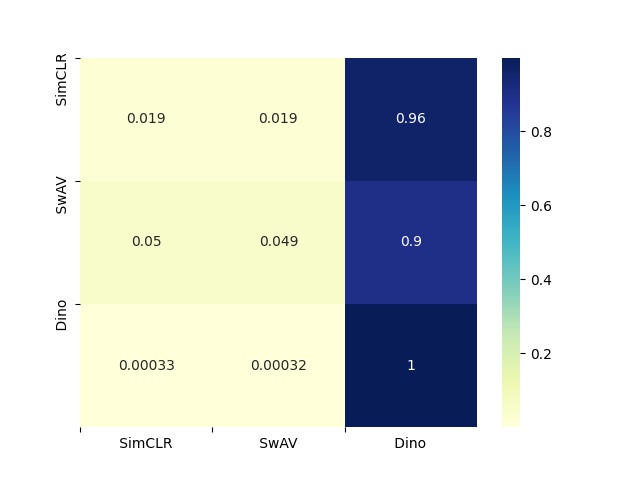}}
         
     \subfigure[Pneumonia Chest X-ray][b]{
        \label{fig:attn-pneucx}
         \includegraphics[width=0.4\textwidth]{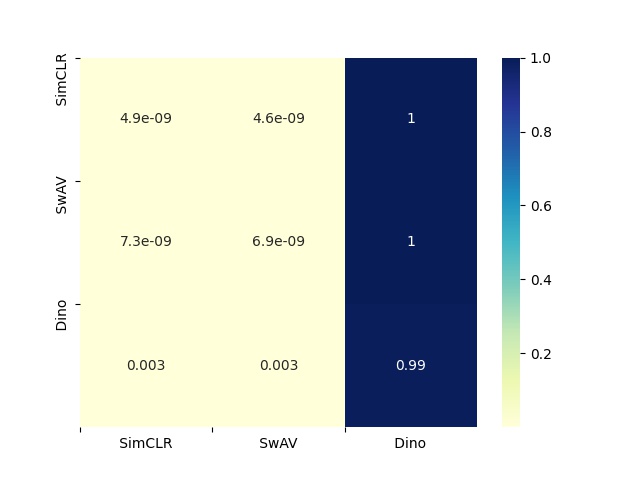}}
     \subfigure[NIH Chest X-ray][b]{
        \label{fig:attn-nih}
         \includegraphics[width=0.4\textwidth]{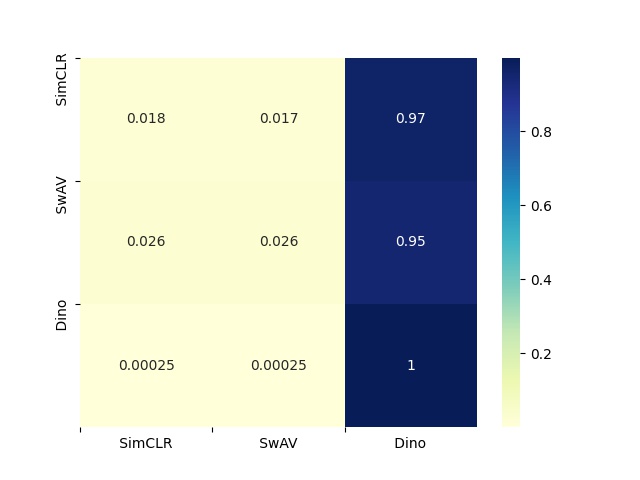}}
    }
\end{figure*}

\section{Number of trainable parameters}
\setcounter{table}{0}
\renewcommand{\thetable}{G\arabic{table}}
\begin{table*}[htbp]
\centering
\floatconts
{table:training-params}
{\caption{
  The number of trainable parameters across all architectures in linear evaluation and finetuning.
  }}
  
  {
\begin{threeparttable}
  \begin{tabular}{c|c|c}
    \toprule
    Architecture & Evaluation Setting & Number of trainable parameters\\    
    \midrule
    \multirow{3}{7em}{ResNet50 \\} 
    & Linear Evaluation   &  10.2 K \\
    & Finetuning & 23.5 M \\
    \midrule
    \multirow{3}{7em}{ViT \\} 
    & Linear Evaluation   &  7.7 K\\
    & Finetuning &  21.7 M\\
    \midrule
    \multirow{3}{7em}{DVME\\} 
    & Linear Evaluation   &  3.6 M\\
    & Finetuning & 72.4 M\\
    \bottomrule
  \end{tabular}

\end{threeparttable}}
\end{table*}
\end{document}